# Portrait Segmentation Using Deep Learning

Sumedh Vilas Datar, Jesus Gonzales Bernal, The University of Texas at Arlington

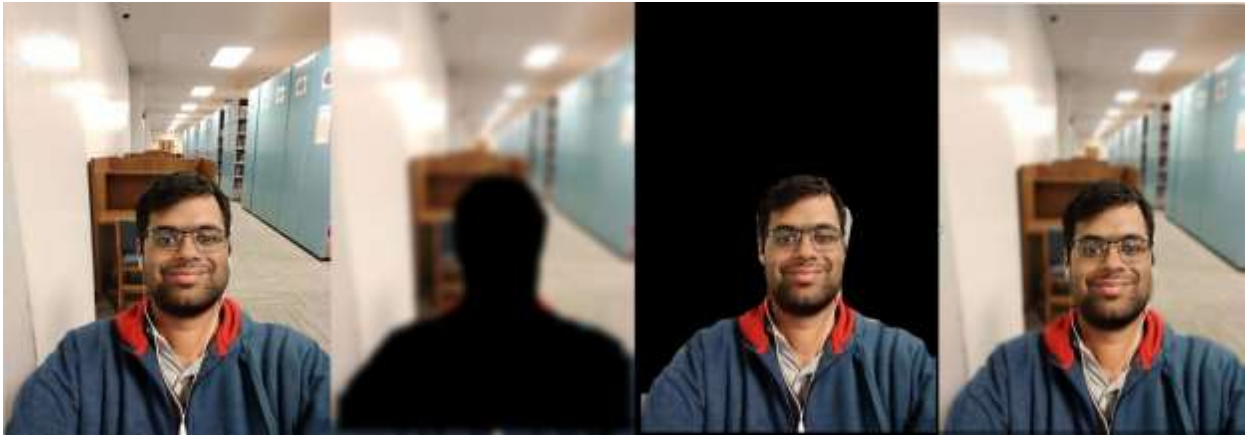

*Figure 1: The stages to create a portrait mode image from a normal image. (a) is the input image (b) Segmenting the foreground and blurring the background (c) Extraction of only foreground (ROI) and masking other regions (d) Final portrait mode image. The image is taken from a normal android phone with a single camera.*

---

**Abstract**

*A portrait is a painting, drawing, photograph, or engraving of a person, especially one depicting only the face or head and shoulders. In the digital world the portrait of a person is captured by having the person as a subject in the image and capturing the image of the person such that the background is blurred. DSLRs generally do it by reducing the aperture to focus on very close regions of interest and automatically blur the background. In this paper I have come up with a novel approach to replicate the portrait mode from DSLR using any smartphone to generate high quality portrait images.*

---

## 1. Introduction

There has been a rise in the number of portrait mode images and a lot of display pictures on Facebook or other social media is being changed to portrait mode and increasing the quality of the image. The main reason is not because of the rise in people buying DSLRs but because of the technology used in DSLR is made available on a smartphone like Google Pixel or Iphone. Also, it is possible to convert a normal image to portrait mode image using an online photo processing application to mark the foreground and blur the background. In this paper I used an approach using deep learning to automate this process and generate a portrait mode image given the image with the subject in the image.

Using techniques like face detection, RCNN and the like cannot be used because we need to segment the region of interest in any shape (on a pixel level) to classify each pixel and not just having a rectangle. In this paper I propose a technique called as image segmentation using a fully convolutional neural network to segment the region of interest and perform foreground extraction and blur the background using a Gaussian Kernel.

## 2. Related Work

The approach used to implement is image segmentation using fully convolutional network (FCN) to extract the foreground. Image blurring to blur the background and image blending to blend the foreground and background images so that you get one single image which is the portrait mode image.

### 2.1 Image Segmentation

Convolutional neural networks have helped in getting near human accuracy for recognition and localization. This is because it has many deep layers enabling the network to learn many features from the image. With huge amount of data and novel approaches to training, the network is getting very accurate to detect region of interest.

I have used a network called MaskRCNN which provides two functionalities which are giving the bounding box and ROI with polygon. I have tweaked the network slightly and removed the functionality of getting bounding box, only giving the output of the region of interest with a polygon using Fully Convolutional Network.

### 2.1a Fully Convolutional Network

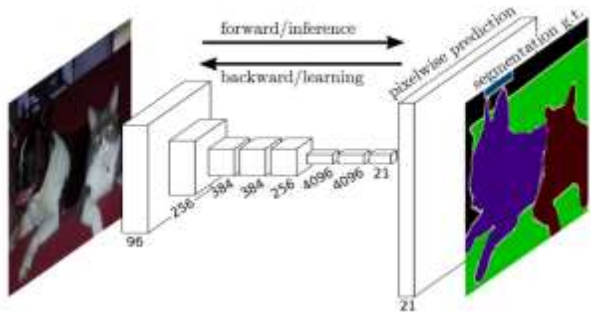

*Figure 2: A fully convolutional neural network*

A fully convolutional network has two parts which are convolution and deconvolution. The convolution layers will help in extracting various features and localizing the region of interest at a smaller dimension and the deconvolution layers will help in projecting the region of interest. Figure 2 shows the diagram of a Fully Convolutional Network.

**Architecture of Fully Convolutional Network.**

Modern semantic image segmentation frameworks are based on the fully convolutional neutral network (FCN) which replaces the fully connected layers of a classification network with convolutional layers. The FCN uses a spatial loss function and is formulated as a pixel regression problem against the ground-truth labeled mask. The objective function can be written as,

$\varepsilon(\theta) = \sum_p e(X\theta(p), \ell(p))$, (1) where p is the pixel index of an image.

$X\theta(p)$ is the FCN regression function in pixel p with parameter $\theta$.

The loss function (softmax) measures the error between the regression output and the ground truth $\ell(p)$. FCNs are typically composed of the following types of layers:

**Convolution Layers** This layer applies several convolution kernels to the previous layer. The convolution kernels are trained to extract important features from the images such as edges, corners or other informative region representations.

**ReLU Layers** The ReLU is a nonlinear activation to the input. The function is $f(x) = \max(0, x)$. This nonlinearity helps the network compute nontrivial solutions on the training data.

**Pooling Layers** These layers compute the max or average value of a particular feature over a region in order to reduce the feature's spatial variance.

**Deconvolution Layers** Deconvolution layers learn kernels to upsample the previous layers. This layer is central in making the output of the network match the size of the input image after previous pooling layers have downsampled the layer size.

**Loss Layer** This layer is used during training to measure the error (Equation 1) between the output of the network and the ground truth. For a segmentation labeling task, the loss layer is computed by the softmax function. Weights for these layers are learned by backpropagation using stochastic gradient descent (SGD) solver.

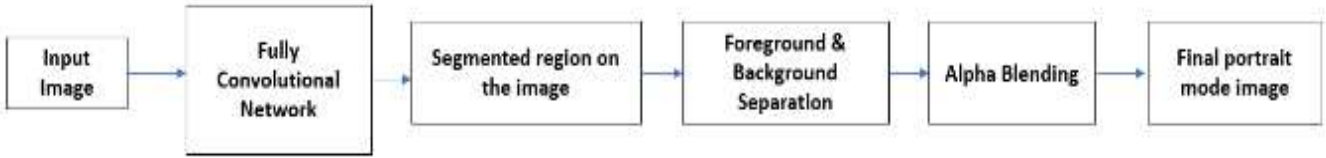

*Fig 3: Working pipeline to generate a portrait mode image*

### 2.1b Image Blending

For image blending I am using a technique called alpha blending which helps in overlaying a foreground image with transparency over a background image. The transparency is often the fourth channel of an image ( e.g. in a transparent PNG), but it can also be a separate image. This transparency mask is often called the alpha mask or the alpha matte.

The math behind alpha blending is straight forward. At every pixel of the image, we need to combine the foreground image color (F) and the background image color (B) using the alpha mask ($\alpha$).

Note: The value of $\alpha$ used in the equation is actually the pixel value in the alpha mask divided by 255. So, in the equation below, $0 \leq \alpha \leq 1$

$$I = \alpha F + (1 - \alpha) B$$

From the equation above, you can make the following observations.

1. When $\alpha = 0$, the output pixel color is the background.
2. When $\alpha = 1$, the output pixel color is simply the foreground.
3. When $0 < \alpha < 1$ the output pixel color is a mix of the background and the foreground. For realistic blending, the boundary of the alpha mask usually has pixels that are between 0 and 1.

### 3. Implementation Details

Image of a person in portrait setting is captured and it is fed through a FCN network. The output from the FCN network is visualized, and it segments the region of interest. The output image with the masked region is used to separate the foreground and background. Figure 4 shows the sample output from the Deep Network.

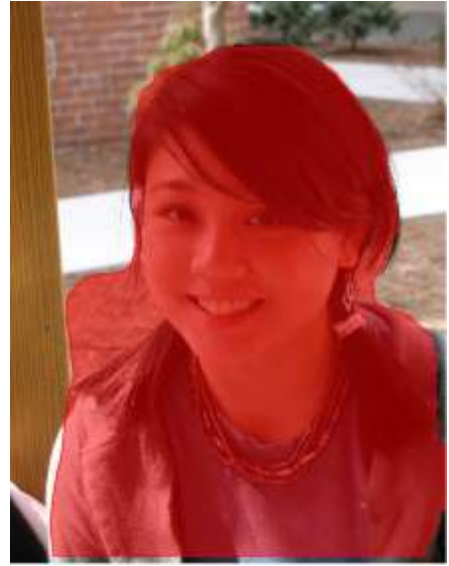

*Fig 4: Output image visualization from the neural network*

The background image is blurred, and then alpha blending is performed on the foreground image which is later overlayed on the background image. This results in creating the portrait mode image. Figure 3 shows the fully working pipeline to generate portrait mode images.

### 4. Dataset and description:

The dataset used is Helen Dataset consists of 2000 training images and 330 test images.

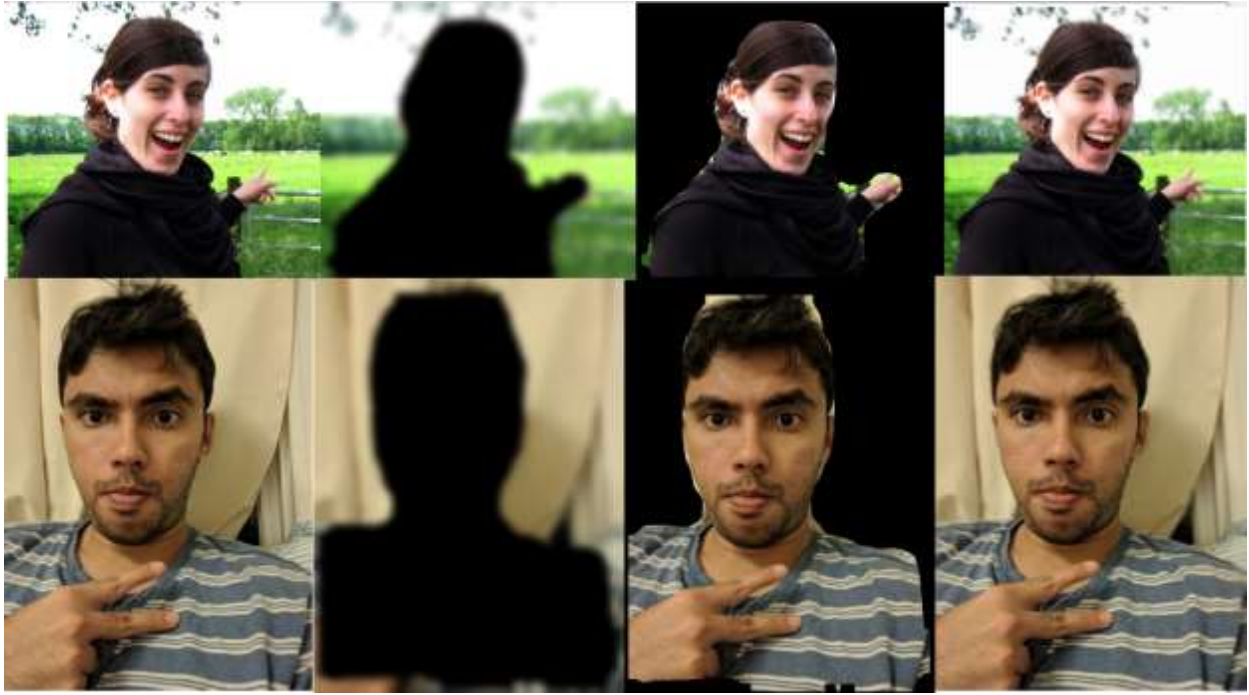

*Fig 5: Showing stages of the pipeline with sample pictures from the Helen Dataset*

Specifically, the dataset was constructed as follows: First, a large set of candidate photos was gathered using a variety of keyword searches on Flickr. In all cases the query included the keyword ``portrait'' and was augmented with different terms such as ``family'', ``outdoor'', ``studio'', ``boy'', ``wedding'', etc. (An attempt was made to avoid cultural bias by repeating the queries in several different languages.) A face detector was run on the resulting candidate set to identify a subset of images that contain sufficiently large faces (greater than 500 pixels in width). The subset was further filtered by hand to remove false positives, profile views, as well as low quality images. For each accepted face, we generated a cropped version of the original image that includes the face and a proportional amount of background. In some cases, the face is very close or in contact with the edge of the original image and is consequently not centered in the cropped image. Also, the cropped image can contain other face instances since many photos contain more than one person in close proximity.

5.  **Training:**

MaskRCNN network was trained on MS COCO dataset and I used the pretrained model to train it on Helen Dataset which consists of over 2000 Images for training and 330 for testing. The loss function used was Softmax and Adam was used as optimizer. The entire application was developed using Python, Keras with Tensorflow backend.

6.  **Results:**

The result for a sample image from helen dataset is shown in Figure 5. Apart from portrait mode the network can be used in various other computer vision problems to perform segmentation techniques.

7.  **Future Scope**

1.  Selective blurring can be used, the network can be trained with multiple objects and many objects in the image can be highlighted and just the background can be blurred. This helps in enhancing the quality of the captured image.

8.  **References**

1.  Automatic Portrait Segmentation for Image Stylization by Xiaoyong Shen1† , Aaron Hertzmann2 , Jiaya Jia1 , Sylvain Paris2 , Brian Price2 , Eli Shechtman2 and Ian Sachs2 1The

Chinese University of Hong Kong 2Adobe Research
2. https://github.com/matterport/Mask_RCNN
Mask R-CNN Kaiming He Georgia Gkioxari Piotr Dollar Ross Girshick ´ Facebook AI Research (FAIR)
3. CS231n by Andrej Karpathy to learn basics of Deep Learning and applications in Computer Vision.
4. https://blog.goodaudience.com/using-convolutional-neural-networks-for-image-segmentation-a-quick-intro-75bd68779225
5. https://blog.goodaudience.com/using-convolutional-neural-networks-for-image-segmentation-a-quick-intro-75bd68779225
6. http://www.ifp.illinois.edu/~vuongle2/helen/
7. https://www.learnopencv.com/alpha-blending-using-opencv-cpp-python/ - Learn OpenCV